\pdfoutput=1

\documentclass[11pt]{article}

\usepackage[preprint]{coling}

\usepackage{times}
\usepackage{latexsym}
\usepackage{booktabs}
\usepackage{amsmath}

\usepackage[T1]{fontenc}

\usepackage[utf8]{inputenc}

\usepackage{microtype}

\usepackage{inconsolata}

\usepackage{graphicx}

\title{RED-CT: A Systems Design Methodology for Using LLM-labeled Data to Train and Deploy Edge Linguistic Classifiers}

\author{
  \textbf{David Farr \textsuperscript{1,2}},
  \textbf{ Nico Manzonelli \textsuperscript{2}},
  \textbf{ Iain Cruickshank \textsuperscript{3}},
  \textbf{ Jevin West\textsuperscript{1}}
\\
\\
\\
  \textsuperscript{1}University of Washington,
  \textsuperscript{2}Army Cyber Technology and Innovation Center,
  \textsuperscript{3}Carnegie Mellon University
\\
  \small{
    \textbf{Correspondence:} \href{mailto:dtfarr@uw.edu}{dtfarr@uw.edu}
  }
}

\begin{document}
\maketitle
\begin{abstract}
Large language models (LLMs) have enhanced our ability to rapidly analyze and classify unstructured natural language data. However, concerns regarding cost, network limitations, and security constraints have posed challenges for their integration into industry processes. In this study, we adopt a systems design approach to employing LLMs as imperfect data annotators for downstream supervised learning tasks, introducing system intervention measures aimed at improving classification performance. Our methodology outperforms LLM-generated labels in six of eight tests and base classifiers in all tests, demonstrating an effective strategy for incorporating LLMs into the design and deployment of specialized, supervised learning models present in many industry use cases.
\end{abstract}

\section{Introduction}

Large Language Models (LLMs) have significantly improved the ability to rapidly evaluate large amounts of unstructured natural language data. Despite their promise, many organizations face internal obstacles integrating LLMs into production environments. Developing LLMs internally is resource, expertise, and time intensive. Likewise, relying on APIs to access external LLMs introduces other issues. For instance, many organizations often have cost constraints, data privacy concerns, air-gapped networks, or decision cycle times that make integrating commercially available APIs infeasible. 

\begin{figure}[ht]
    \centering
    \includegraphics[scale=.275]{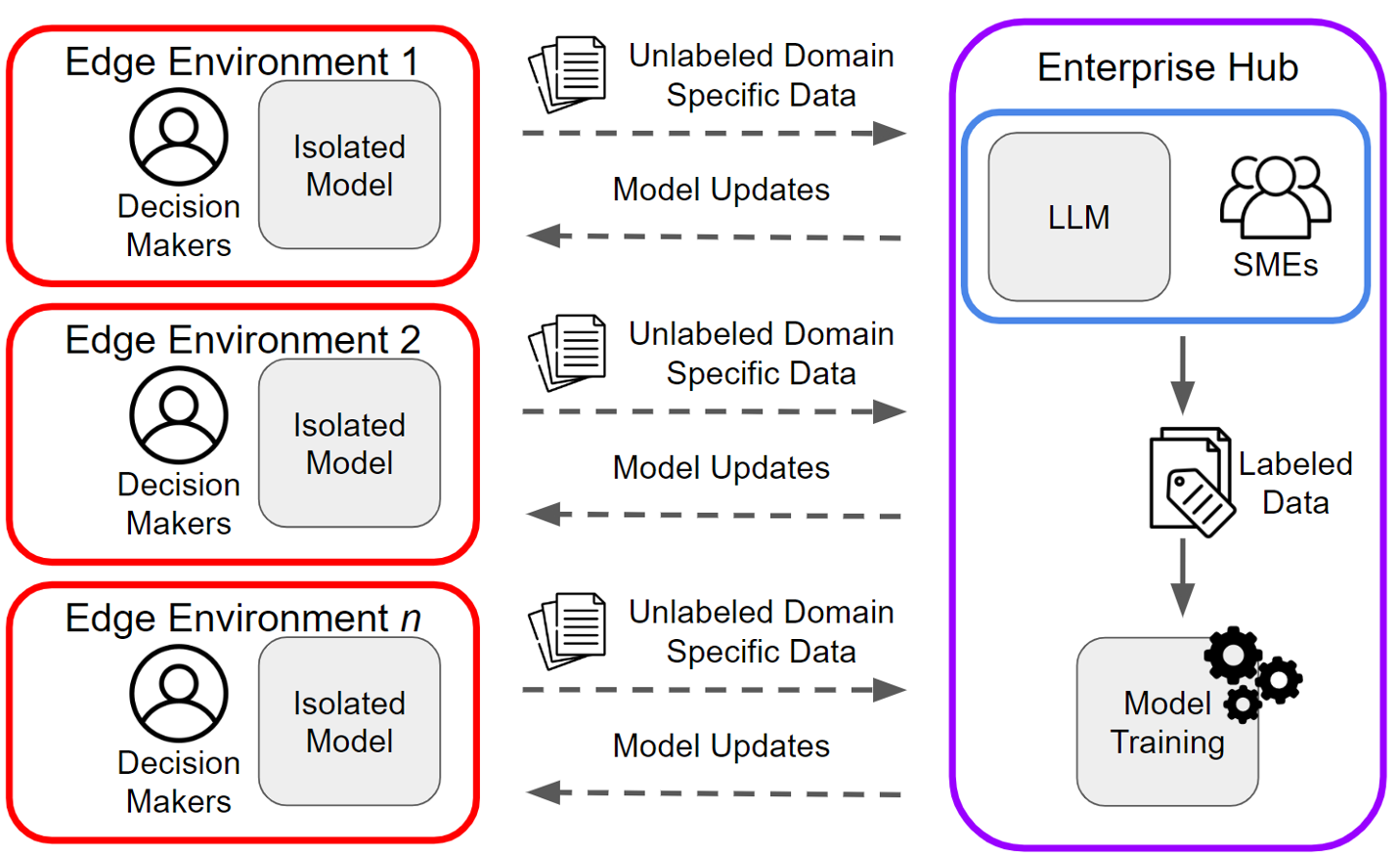}
    \caption{RED-CT design which allows LLM-like capabilities for NLP tasks deployed in edge environments.}
    \label{fig:RES2D}
\end{figure}

Prior work shows that LLMs can perform well across a variety of NLP tasks for computational social science (CSS) via zero-shot prompting \citep{ziems24}. Traditionally, these tasks, like emotion, stance, persuasion, and misinformation classification, are solved with classification via supervised learning techniques. Although using supervised models solves many issues associated with deploying LLMs in production environments, they are known to perform poorly on out-of-domain data and require a significant upfront investment in data labeling.

To balance the flexibility associated with LLMs and advantages of supervised models, we propose Rapid Edge Deployment for CSS Tasks (RED-CT). RED-CT is a system that integrates traditional techniques from active learning such as confidence measurements and soft labels to pair LLM generated data labels with minimal selected human-annotated labels to deploy classifiers to edge environments fast. 
We define the \textit{edge environment} as time- and / or resource-limited situations where users need to interface with NLP solutions. Additionally, the edge environment may be disconnected from the internet for security or privacy purposes or in crisis response settings where connection to the internet is either impracticable or unreliable.

In this paper, we introduce RED-CT and propose a confidence-informed sampling method to select LLM-labeled data for human annotation. In addition, we present a simple method to generate soft labels from LLM predictions to use during edge classifier training. We evaluate RED-CT with confidence-informed sampling and learning on soft labels across four CSS tasks: stance detection, misinformation identification, humor detection, and ideology detection. We further evaluate the proposed approach with two different common data-labeling prompting schemes and across three different sizes of distilled models. Our results show that it is possible to approximate or outperform LLMs on CSS tasks with minimal human data labeling ($~10\%$ of dataset) in the distillation of edge models.

\section{Related Works}

One of the chief issues in creating ML solutions for CSS tasks is generalizing to out-of-domain data. CSS tasks, such as stance detection or sarcasm classification, often have very nuanced, context-dependent language \cite{ngcarley, ziems24, cruickshank2024prompting}. Due to high contextually-dependency, supervised approaches produce models that struggle to generalize between datasets. For example, previous research indicates that while model generalizability can be improved through the aggregation of datasets, cross-dataset stance detection models still generalize poorly \cite{ngcarley}.

Recent work has demonstrated that LLMs can perform well across various classification tasks within CSS \cite{cruickshank2024prompting, zhu2023chatgpt}. \citet{ziems24} provides best practices for prompting and benchmarks performance for a variety of CSS tasks across several LLMs. LLM-based classification methods work better with out-of-domain data due to the LLMs strong zero-shot classification capacity. However, these methods also require substantial resources and cannot scale, in terms of cost or compute time, to large CSS datasets. For example, just labeling the SemEval2016 dataset \cite{StanceSemEval2016} (2,814 data points) with GPT-4 could cost over \$30 USD. Additionally, ongoing research has found that LLMs still usually perform worse than in-domain supervised models at CSS tasks \cite{cruickshank2024prompting, ziems24}. \cite{tan2024largelanguagemodelsdata} provide a survey paper of research using large language models for data annotation, including model distillation as a task. Some related works included show using synthetic generated data from larger LLMs to train smaller LLMs\cite{wang2023letssynthesizestepstep} and \cite{huang2022largelanguagemodelsselfimprove} which demonstrate LLMs can improve performance through self-annotation and subsequent fine-tuning based on self-annotated data. Further related work by \cite{human-llm} deploys an external verifier model to select samples LLMs are unlikely to classify correctly and routes them to human labelers for increased performance. Such previous work differs in data sampling methods, resource requirements, and distillation methodology.

In an effort to improve supervised model performance in other classification contexts, researchers have explored learning on soft labels. Soft labels employ a weighting mechanism to capture annotator uncertainty during labeling. Soft labels have been shown to enhance model generalization and better represent the confidence of the annotator \cite{softlabels, wu-etal-2023-dont}.

Researchers study LLM distillation techniques \cite{knwoledgedist} to reduce model size and cost. These methods vary considerably in their use of LLMs. Some studies have focused on generating artificial data with LLMs useful for distilling small classification models \cite{ye2022zerogen, ye2022progen, gao2022self, meng2023tuning}. Other works have explored few-shot prompting and active learning mechanisms, combined with LLMs for data labeling \cite{wang-etal-2021-want-reduce, zhang2023llmaaa, hsieh2023distilling}. Many of these methods often require human intervention to filter low-quality data or LLM-generated rationales for labels which can be unreliable \cite{huang2023can}. Other works focus on reducing bias \cite{SurrogateLabels} without focusing on downstream classification performance \cite{wang-etal-2021-want-reduce}. \citet{supervisedllm} assess supervised classifiers performance on LLM generated labels, but do not offer a systems approach or intervention measures to improve downstream classification. None of the prior works attempt to integrate additional uncertainty information from LLMs into human intervention and model distillation.

\section{Methodology}

In this section, we outline our proposed methodology that contributes to the literature by presenting a systems approach that incorporates model uncertainty estimates. These estimates guide human intervention and improve model training for classifiers using LLM-labeled data.

\subsection{RED-CT System Methodology}

Rapid Edge Deployment for CSS Tasks (RED-CT) is designed with three tasks in mind: reducing latency for classification tasks, reducing the amount of data exposed to external API's, and decreasing the energy and monetary cost associated with LLMs. By reducing LLM dependency, we can decrease energy expenditure, cost, and network dependency for CSS classification tasks. This also allows us to obfuscate batched data being sent to an LLM, opposed to needing to secure all data in a production environment. RED-CT is a system that enables users in edge environments to utilize ML tools for complex societal computing tasks. Figure 1 provides a high-level overview of our system.

RED-CT follows a framework in which classification and data collection are performed at the edge, model development is performed at a central point, and then model updates are pushed back to the edge. We refer to this framework of different, related contexts and devices as a data resupply framework. Transport mechanisms for data resupply include internet (when available) or physical devices transferred by personnel moving in and out of the edge environment.

Data delivered to the enterprise hub goes through a pipeline for labeling and model training. Unlike the edge environment, compute resources and connectivity are not restricted at the enterprise hub. This allows analysts at the hub to label the data via zero-shot LLM prediction for maximum expediency. Data label quality can be increased by integrating subject matter experts (SMEs) for prompt engineering, quality control, or expert labeling of small sample sizes. Edge classifiers are then trained or fine-tuned on the newly labeled data and deployed back to the edge environment.

RED-CT's modular design allows for increased performance as industry and academia continue to improve system components, such as LLMs, prompting techniques, and edge classifiers. Additionally, our method prevents model drift by enabling constant evaluation of data in a dynamic environment, with human-in-the-loop processes informing users.

\subsection{Training Edge Classifiers on LLM-labeled Data}

Due to the potential time-constrained setting in edge environments, RED-CT relies on fine-tuning BERT-based models on LLM-labeled data. BERT-based models require minimal text preprocessing, and their pretraining allows for fine-tuning on downstream tasks. BERT models exhibit strong performance when fine-tuned for a variety of classification tasks \cite{devlin2019bert}.

Given that LLMs are prone to errors in the zero-shot prediction setting, we assume that our LLM labels will be imperfect. Naively fine-tuning BERT on the LLM-labeled data risks over fitting to noisy or incorrect labels. To improve edge model performance, we integrate several system interventions into the model fine-tuning process: including expert-labeled data into the training process, designing confidence scores to select samples for experts to label, and learning soft labels based on label weights.

\subsubsection{Incorporating Confidence Informed Expert Labels}

RED-CT helps streamline model deployment by reducing the number of personnel hours devoted to labeling data. Instead of using SMEs to label all available data, we only require them to label small samples of data. Integrating experts improves the quality of the LLM-labeled dataset and subsequently the edge classifier.

Randomly selecting samples for SME labeling within a bounded time or up to a certain percentage can improve edge model performance but may introduce inefficiencies where SME's analyze sample data in which the LLM is confident it has labeled correctly. To optimize sampling for SME analysis, we devise a confidence-based metric to identify examples where LLM labeling is less reliable.

The confidence score is defined as the absolute difference between the highest token label log probability and the second-highest token label log probability within this constrained set of expected tokens. Let \(\mathcal{T}\) represent the set of given tokens, and \(P(t)\) denote the distribution of log probabilities across each token \(t \in \mathcal{T}\). The confidence score, denoted as \(C\), is then computed using the formula

\begin{equation}
    \text{\(C\)} = \left| \max_{t \in \mathcal{T}} P(t) - \max_{t \in \mathcal{T} \setminus \{t^*\}} P(t) \right|,
\end{equation}

\noindent where \(t^*\) is the token corresponding to the highest probability \( \max_{t \in \mathcal{T}} P(t) \). To apply the confidence score, we stratify by each LLM-labeled class and sample the bottom $p$ percentile.

\begin{figure}
    \includegraphics[scale=.60]{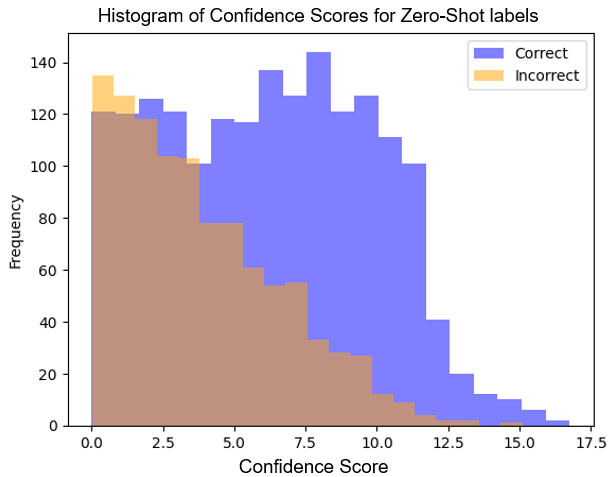}
    \caption{The distribution of confidence scores for examples labeled correctly and incorrectly using gpt-3.5-turbo zero-shot stance classification. The distributions are overlaid as opposed to stacked.}
    \label{fig:cscore}
\end{figure}

To validate the proposed confidence estimate, we analyze the distribution of confidence scores for examples labeled correctly and incorrectly using the labels from zero-shot stance classification with gpt-3.5-turbo. Under the Kolmogorov-Smirnov test, we reject the null hypothesis that correctly and incorrectly labeled samples come from the same distribution of confidence scores \cite{an1933sulla}. Highlighted in Figure \ref{fig:cscore}, we are more likely to select correctly labeled examples using random sampling, but less likely when sampling examples with very low confidence scores.

\subsubsection{Learning on Soft Labels}

Fine-tuning edge classifiers on the LLM-labeled data risks overfitting on incorrect labels. We help ease this problem by integrating SMEs into the labeling process; however, standard supervised training methods do not account for differences in label quality. To account for label confidence, we learn on \textit{soft labels}.

To retrieve a soft label for model fine-tuning, we apply the expit function to the probability of the token associated with each LLM label. For expert labeled examples, we assign a weight of 1 on the selected class and 0 for the others. Our experimental results show that learning with soft labels improves edge classification performance.

\begin{table*}[ht]
    {
    \centering
    \begin{center}
    \begin{tabular}{@{}l|l|llll@{}}
    \toprule
    Task & \multicolumn{1}{c|}{Enterprise LLM} & \multicolumn{3}{c}{Edge Classifier - RoBERTa-L} \\
    \bottomrule
    \bottomrule
    & \multicolumn{1}{c|}{GPT-3.5 Turbo} & \multicolumn{1}{c}{Random} & \multicolumn{1}{c}{Base} & \multicolumn{1}{c}{RS 10\%}  & \multicolumn{1}{c}{CI SL 10\%} \\ 
    \midrule
    Stance & .667 & .333 &.626  & .665 &  \textbf{.689} \\
    Misinformation & .761 & .500 & .653 &.703  & .752 \\
    Ideology & .579 & .333 & .567  & .597 & \textbf{.626}  \\
    Humor & .565 & .500 & .534  & .555 &  \textbf{.571} \\
    \bottomrule
     & Mistral-7B-Instruct & & & \\
     \bottomrule
    Stance & .529 & .333 & .439 & .448 & .486 \\
    Misinformation & .602 & .500 & .594 & .629  & \textbf{.665} \\
    Ideology & .406 & .333 & .413 & .441 & \textbf{.451} \\
    Humor & .492 & .500 & .384 & .427 & \textbf{.508} \\
    \bottomrule
    \end{tabular}
    \end{center}
    }

    \caption{Zero-Shot LLM performance (weighted f1 score) compared to edge model performance. Random are dummy models predicting on a uniformed distribution, base edge models are trained without system interventions, RS 10\% edge models are trained with 10\% randomly sampled expert examples, and CI SW 10\% is 10\% confidence-informed sampling and learning with label weights. Results that out-performed the enterprise LLM are bolded.
    \label{tab:css-results}}
\end{table*}

\section{System Implementation and Experiment Design}

We replicate the available LLM labeling capabilities with two models: OpenAI's gpt-3.5-turbo, available closed-source from their API, and Mistral's Mistral-7B-Instruct-v0.2, available open-source on Huggingface.\footnote{Model resources and information are contained in the Ethics and Availability section.} We experiment with two different prompting styles for labeling: zero-shot and zero-shot chain of thought (CoT). We attempted to use the best prompting practices in literature for our classification tasks, integrating prompting techniques from \cite{ziems24}, \cite{cruickshank2024prompting}, and \cite{zhu2023chatgpt}. Examples of each prompt are provided in Appendix \ref{sec:appendix-prompts}.

For edge classifiers, we test three flavors of BERT: 'Distil-BERT', 'RoBERTa', and `RoBERTa-Large' \cite{devlin2019bert, liu2019robertarobustlyoptimizedbert}. These models vary in size, allowing us to assess performance across model compute requirements. For each BERT model, we evaluate the effects of system intervention measures. The system settings we tested included a base classifier trained with no system interventions directly on the LLM labels, a classifier trained on soft labels (SL), a classifier trained on 10 percent randomly selected expert labeled data (RS 10\%), a classifier trained on confidence-informed sampling (CI 10\%), and a classifier trained with all system intervention measures (CI SL 10\%). We train five classifiers on each LLM-labeled dataset and report the averages across each. For each edge model, we do full fine-tuning (i.e., unfreeze all model weights) from pre-trained models, but note that this process can be done with any type of fine-tuning or training a model with initialized weights.

\subsection{CSS Tasks and Data Selection}

For the purposes of testing our systems methodology, we selected four well known CSS tasks: stance detection, misinformation detection, ideology detection, and humor detection. We then selected a dataset for each task that had known benchmarks to compare our system design against.

\subsubsection{Stance Detection}

We define stance detection as an "automatic classification of the stance of the producer of a piece of text, towards a target, into one of these three classes: {Favor, Against, Neither}" \cite{stancedef}.  We use the SemEval-16 dataset provided by \cite{StanceSemEval2016}. The SemEval-16 dataset consists of approximately 5000 tweets in relation to one of five targets: Hilary Clinton, Legalization of Abortion, Feminism, Climate Change, and Atheism. There are three classification classes for each target: favor, against, and neutral.

\subsubsection{Misinformation}

We define misinformation as "false or inaccurate information that is deliberately created and is intentionally or
unintentionally propagated" \cite{misinfodefined}.
We evaluate misinformation detection on the Misinfo Reaction Frames corpus \cite{misinfodata}. The Misinfo Reaction Frames corpus consists of 25k news headlines consisting of topics such as COVID-19, climate change, or cancer. Each headline was fact checked and has an associated binary misinfo classification of misinformation or trustworthy.

\subsubsection{Ideology}

We define ideology as "the shared framework of mental models that groups of individuals possess that provide both an interpretation of the environment and a prescription as to how that environment should be structured" \cite{ideologydefined}. We used the Ideology Books Corpus (IBC) dataset from \cite{sim-etal-2013-measuring} with sub-sentential annotations \cite{iyyer2014neural} to evaluate our system's utility in ideology detection. The IBC dataset contains 1,701 conservative sentences, 600 neutral sentences, and 2,025 liberal sentences.

\subsubsection{Humor}
For humor detection, we used a broad definition when prompting LLMs with the question, "Would most people find this funny?" This approach focused on binary humor classification. We evaluated our system using a curated collection of posts from Reddit's r/Jokes, where researchers labeled jokes as humorous or not based on the number of upvotes. The two classes were distinguished through binary cluster analysis \cite{weller-seppi-2019-humor}. 

\section{Results}

\begin{figure*}[h]
    \centering
    \includegraphics[scale=.50]{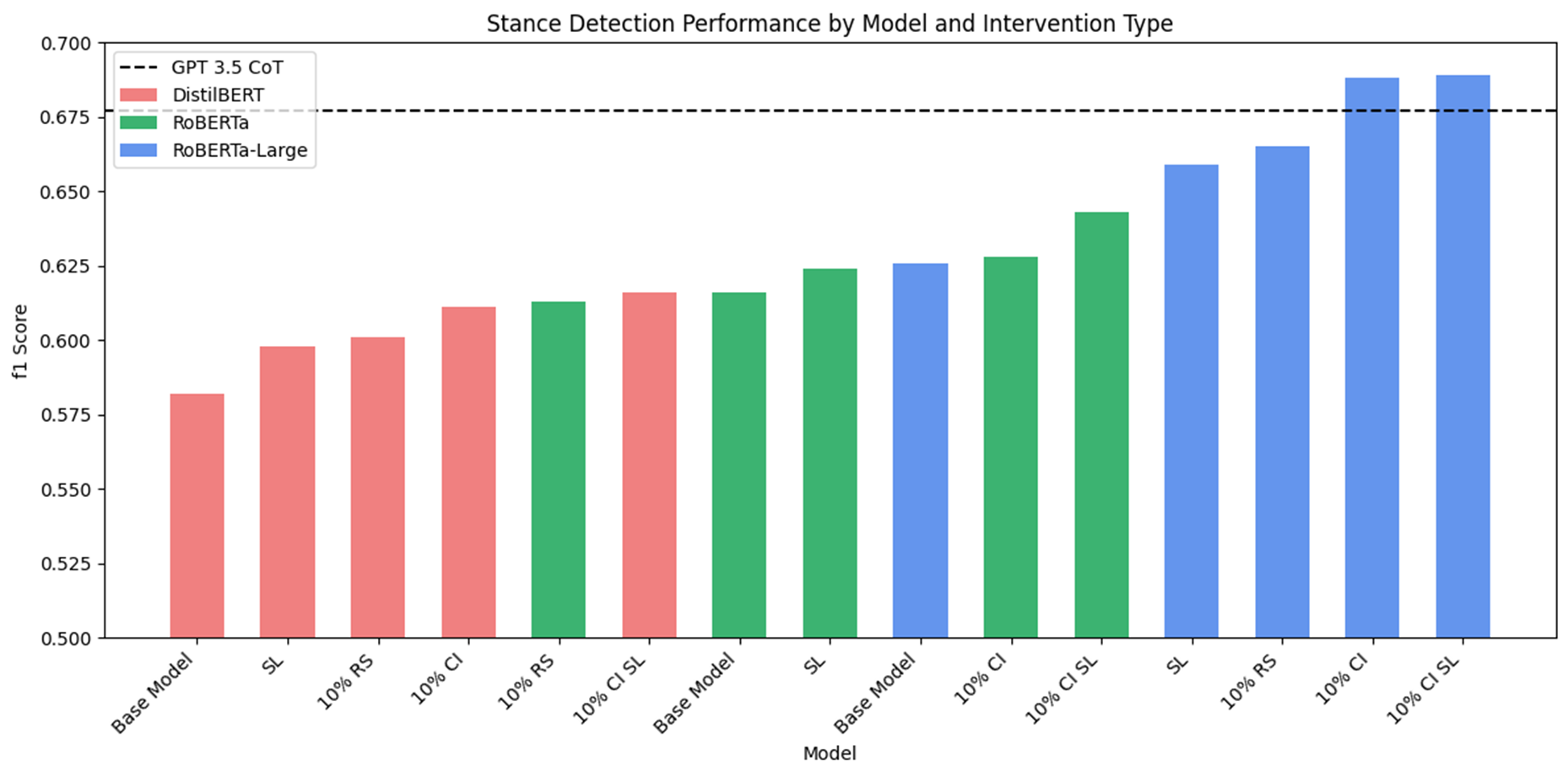}
    \caption{Comparing edge model F1 score as we change model and system interventions types for stance detection. We note steady improvements of edge model performance as we introduce more complex models and system intervention measures. The largest edge model with all system interventions out-performs gpt-3.5-turbo CoT.}
    \label{fig:semcot}
\end{figure*}

Table \ref{tab:css-results} presents the high-level results across our four chosen CSS tasks using RoBERTa-L. Figure \ref{fig:semcot} is a more detailed analysis of the implementation of our system methodology in the stance detection task, including varying the type of BERT model in all combinations of system intervention strategies. A key takeaway is that through our methodology and associated system intervention measures, we were able to outperform LLM-labeled data in 6 of the 8 tested tasks, while approximating it in an additional task. Additionally, in GPT labeled data, we had an average improvement of 6.75\% over the base classifier and we out performed the base and normal sampling techniques in 100 percent of tasks. In Annex B, we have included additional results analysis, including varying the percentage of expert labels in Figure \ref{fig:vary_p} and a full table for each stance detection result in Table \ref{tab:sem-eval-results}.

\subsection{Discussion}

Our results represent a significant improvement in system design for using LLMs as imperfect annotators for downstream classification tasks. Our system intervention measures were effective in both GPT 3.5 and Mistral-7B, but more consistent in GPT 3.5. We theorize that this is because the logits returned from GPT 3.5 provided more value to our confidence score and weighting interventions due to better associated logit values with correct classification. Furthermore, we noticed some bias in LLM classification where the LLM was consistently incorrect in predicting a single class. This was represented in our confidence scores and caused our expert labels to focus on a single class, resulting in heavily weighted soft labels applied to a single class extrapolating existing error. To solve this problem, we stratified the expert sampling process, selecting the bottom 10 percent of confidence scores for each class instead of the bottom ten percent of the entire dataset. Doing so slightly decreased the accuracy on tasks where there was minimal bias, but greatly increased the accuracy where LLM bias was present such as ideology and stance classification. This difference in class performance for a given task has also been observed in other works. For example, LLMs consistently exhibit a discernible left and libertarian bias, as assessed by political orientation surveys, that likely arises due to the training data used for training LLMs \cite{motoki2023more,rozado2023political,rutinowski2023self}. This bias could affect performance on frequently politically charged tasks (which are also frequently important tasks for CSS), such as stance classification. 

Confidence-informed sampling allowed us to greatly improve our edge classifier and should be integrated into any knowledge distillation process where small batch labeling is incorporated. Our confidence score distributions were the most discernible when ensembling different prompting techniques or in zero-shot settings. Chain-of-thought prompting resulted in less clean distributions, but further testing is required to fully understand the causation of prompting mechanisms on returned logit distributions. 

\section{Conclusions}

In this work, we successfully replicated LLM performance in an edge environment on computational social science tasks using a systems methodology. Our approach, which integrates expert-in-the-loop data labeling for a small portion of the data (10\% or less), enables the deployment of highly performant small models in environments where LLM access is restricted due to cost, security, or latency concerns.

Our results demonstrate generalizability across various labeling prompts and distilled models, providing a flexible and scalable solution. This methodology offers a practical mechanism to reduce labeling costs and dependence on large LLMs while improving performance and data annotation throughput, even in resource-constrained settings with minimal human intervention.

\newpage
\bibliography{custom}

\appendix

\section{Sample Prompts}
\label{sec:appendix-prompts}

This appendix contains example prompts using the same document in the SemEval-2016 dataset. The newlines in the example prompts correspond to actual newline characters ("\textbackslash n") in the prompt string.

\subsection{Stance Detection Zero-Shot}

\begin{quote}
THIS IS AN EXAMPLE STATEMENT\newline
Which of the following best describes the above social media statements' stance regarding TOPIC?\newline
A) For\newline
B) Against\newline
C) Neutral\newline
Only respond with 'For', 'Against', or 'Neutral'. 

\end{quote}

\subsection{Stance Detection Zero-Shot CoT}

\begin{quote}
Stance classification is the task of determining the expressed or implied opinion, or stance, of a statement toward a specific target. Think step-by-step and explain the stance (For, Against, or Neutral) of the following social media statement towards TOPIC.\newline
target: TOPIC\newline
statement: THIS IS AN EXAMPLE STATEMENT\newline
explanation:
\newline
LLM Replies!
\newline
Therefore, based on your explanation, what is the stance of the following social media statement toward the target?
\newline
target: TOPIC\newline
statement: THIS IS AN EXAMPLE STATEMENT\newline
A) For\newline
B) Against\newline
C) Neutral\newline
Only respond with 'For', 'Against', or 'Neutral'. If the statement is not relevant to TOPIC, select Neutral.
\end{quote}

\subsection{Misinformation}
"This is an example headline" \newline
Which of the following describes the above news headline?\newline
A) Misinformation\newline
B) Trustworthy\newline
Only respond with 'Misinformation' or 'Trustworthy'

\subsection{Ideology}
statement: "This is an example ideological statement."\newline
Which of the following leanings would a political scientist say that the above statement has?\newline
A: Conservative\newline
B: Neutral\newline
C: Liberal \newline
Only respond with 'Conservative', 'Neutral', or 'Liberal'

\subsection{Humor}
Joke: This is an example joke.\newline
Would most people find the above joke humorous? You must pick between 'True' or 'False'.\newline You cannot use any words other than 'True' or 'False'.

\section{Stance Classification Additional Results}

\begin{figure*}
    \centering
    \includegraphics[scale=.4]{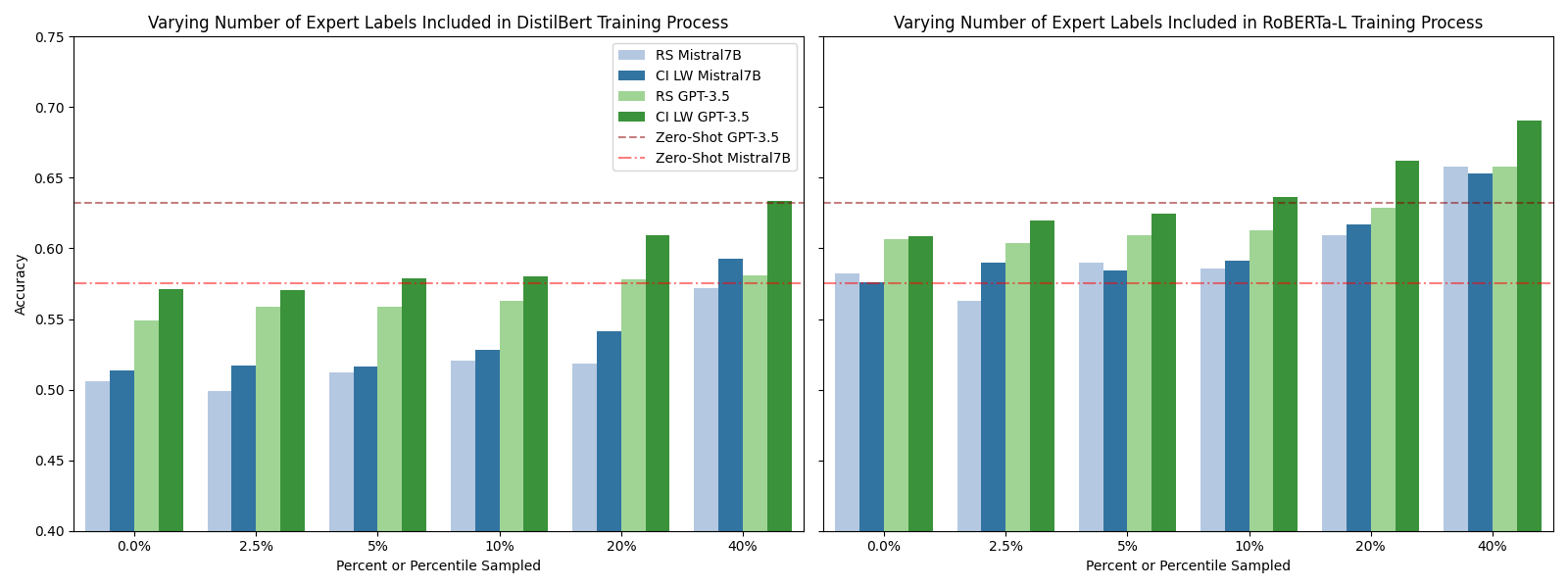}
    \caption{Varying the number of expert labels included amongst the LLM labels in the training process for DistilBERT and RoBERTa-L. RS implies randomly sampled expert labels for the training process and CI SL implies confidence informed sampling with label weighted training. Blue corresponds to the Mistral-7B-Instruct-2.0 LLM labeler and green corresponds to the GPT-3.5 LLM labeler. The horizontal dashed lines represent the zero-shot accuracy of each LLM.}
    \label{fig:vary_p}
\end{figure*}

The appendix contains additional experimental results for the stance detection task. Figure \ref{fig:semcot} shows the effect of different system intervention strategies across the three main flavors of BERT evaluated. Figure \ref{fig:vary_p} shows the effect of varying the percentage of expert labels with various system interventions, and Table \ref{tab:sem-eval-results} contains a table of F1 scores across all interventions.

\begin{table*}
    \centering
    \begin{tabular}{@{}l|l|lllll@{}}
    \toprule
    Prompt Technique & \multicolumn{1}{c|}{Enterprise LLM} & \multicolumn{3}{c}{Edge Classifier - DistilBERT} \\
    \bottomrule
    \bottomrule
    & \multicolumn{1}{c|}{GTP-3.5} & \multicolumn{1}{c}{Base} & \multicolumn{1}{c}{SL} & \multicolumn{1}{c}{RS 10\%} & \multicolumn{1}{c}{CI 10\%} & \multicolumn{1}{c}{CI SL 10\%}\\ 
    \midrule
    \textit{Zero-Shot} & .629 & .549 & .562 & .559 & .570 & .582 \\
    \textit{Zero-Shot CoT} & .677 & .582 & .598  & .601 & .611 & .616 \\ 
    \bottomrule
     & Mistral-7B-Instruct & & & & & \\
     \bottomrule
    \textit{Zero-Shot} & .599 & .485 & .536 & .534  & .536 & .552 \\
    \textit{Zero-Shot CoT} & .589 & .493 & .452  & .519 & .496 & .505 \\ 
    \bottomrule
    \end{tabular}
    \begin{tabular}{@{}l|l|lllll@{}}
    \toprule
    Prompt Technique & \multicolumn{1}{c|}{Enterprise LLM} & \multicolumn{3}{c}{Edge Classifier - RoBERTa} \\
    \bottomrule
    \bottomrule
    & \multicolumn{1}{c|}{GTP-3.5} & \multicolumn{1}{c}{Base} & \multicolumn{1}{c}{SL} & \multicolumn{1}{c}{RS 10\%} & \multicolumn{1}{c}{CI 10\%} & \multicolumn{1}{c}{CI SL 10\%}\\ 
    \midrule
    \textit{Zero-Shot} & .629 & .575 & .587 & .580  & .594 & .615 \\
    \textit{Zero-Shot CoT} & .677 & .616 & .624  & .613 & .628 & .643 \\ 
    \bottomrule
     & Mistral-7B-Instruct & & & & & \\
     \bottomrule
    \textit{Zero-Shot} & .599 & .549 & .539 & .589  & .588 & .565 \\
    \textit{Zero-Shot CoT} & .589 & .530 & .476  & .561 & .554 & .532 \\ 
    \bottomrule
    \end{tabular}
    \begin{tabular}{@{}l|l|lllll@{}}
    \toprule
    Prompt Technique & \multicolumn{1}{c|}{Enterprise LLM} & \multicolumn{3}{c}{Edge Classifier - RoBERTa-L} \\
    \bottomrule
    \bottomrule
    & \multicolumn{1}{c|}{GTP-3.5} & \multicolumn{1}{c}{Base} & \multicolumn{1}{c}{SL} & \multicolumn{1}{c}{RS 10\%} & \multicolumn{1}{c}{CI 10\%} & \multicolumn{1}{c}{CI SL 10\%}\\ 
    \midrule
    \textit{Zero-Shot} & .629 & .603 & .612 & .617 & .618 & \textbf{.637} \\
    \textit{Zero-Shot CoT} & .677 & .626 & .659  & .665 & .\textbf{688} & \textbf{.689} \\
    \bottomrule
     & Mistral-7B-Instruct & & & & & \\
     \bottomrule
    \textit{Zero-Shot} & .599 & .578 & .596 & \textbf{.608}  & \textbf{.613} & \textbf{.610} \\
    \textit{Zero-Shot CoT} & .589 & .\textbf{597}  & .560  & \textbf{.603} & .559 & \textbf{.597} \\ 
    \bottomrule
    \end{tabular}
    \caption{F1 scores on SemEval2016. Edge classifier variants: 'Base' trained on LLM labels directly, SL trained with label weighting, RS 10\% trained with 10\% randomly sampled expert labels, CI 10\% trained with 10\% confidence informed expert labels, and CI SL 10\% trained with 10\% confidence informed expert labels and labeling weighting.}
    \label{tab:sem-eval-results}
\end{table*}

\end{document}